\documentclass[11pt]{article}
\usepackage{colacl, times, myexamples,tipa,amsmath}
\usepackage[dvips]{epsfig}

\newcommand{\ling}[1]{\emph{#1}}
\def\sref#1{\S\ref{#1}}
\def\code#1{{\small\tt #1}}
\let\l\code
\newcommand{\strich}{\rule{.99\linewidth}{.1pt}\\}
\newcommand{\startpiece}{\par\noindent\strich\vspace{-1.1\baselineskip}\footnotesize}
\newcommand{\stoppiece}{\vspace{-\baselineskip}\noindent\strich\par\noindent\normalsize}
\newcommand{\finalpagebreak}{\pagebreak}
\newcommand{\finalforcedpage}{\enlargethispage*{100cm}}

\title{Finite-State Reduplication in One-Level Prosodic Morphology}
\author{Markus Walther \\ University of Marburg \\
            FB09/IGS, Wilhelm-R{\"o}pke-Str. 6A, D-35032 Marburg, Germany \\
           \tt Markus.Walther@mailer.uni-marburg.de}

\begin{document}

\maketitle
\begin{abstract}
Reduplication, a central
instance of prosodic morphology, is particularly challenging for state-of-the-art
computational morphology, since it involves copying of some part of a phonological string.  
 In this paper I advocate a finite-state method that combines enriched 
 lexical representations via intersection to implement the copying.
The proposal includes a resource-conscious variant of automata
and can benefit from the existence of lazy algorithms. Finally, the
implementation of a complex case from Koasati is presented.
\end{abstract}
\section{Introduction}
In the past two decades computational morphology has been quite successful 
in dealing with the challenges posed by natural language word
patterns. Using finite-state methods, it has been possible to  
describe both word formation and the concomitant phonological modifications 
in many languages, ranging from straightforward
concatenative combination \cite{koskenniemi:83} over Semitic-style nonconcatenative
intercalation (Beesley \shortcite{beesley:96}, Kiraz
\shortcite{kiraz:94}) to circumfixional long-distance dependencies
\cite{beesley:98}.

However, Sproat \shortcite{sproat:92} observes that, despite the existence of ``working
systems that are capable of doing a great deal of morphological
analysis'', ``there are still outstanding problems and areas which
have not received much serious attention'' (ibid., 123). Problem areas
in his view include subtractive morphology, infixation, the proper inclusion
of prosodic structure and, in particular, reduplication: ``From a
computational point of view, one point cannot be overstressed: 
the copying required in reduplication places reduplication in a class
apart from all other morphology.'' (ibid., 60). Productive reduplication is so
troublesome for a formal account based on regular languages (or
regular relations) because unbounded total instances like Indonesian noun plural
(\ling{orang-orang} `men') are isomorphic to the copy language
$ww$, which is context-sensitive.

In the rest of this paper I will lay out a proposal for handling
reduplication with finite-state methods. As a starting point, I adopt
Bird \& Ellison \shortcite{bird.ellison:94}'s One-Level Phonology%
, a mono\-stratal constraint-based framework where phonological representations,
morphemes and generalizations are all finite-state automata (FSAs) and constraint
combination is accomplished via automata intersection. While it is possible to
transfer much of the present proposal to the transducer-based setting that is 
often preferred nowadays, the monostratal approach  still offers an attractive alternative
due to its easy blend with monostratal grammars such as HPSG and the
good prospects for machine learning of its surface-true constraints
(Ellison \shortcite{ellison:92}, Belz \shortcite{belz:98}). 

After a brief survey of
important kinds of reduplication in \sref{red}, section \sref{fsm} explains the necessary
extensions of One-Level Phonology to deal with the challenges presented by
reduplication, within the larger domain of prosodic morphology in
general. A worked-out example from Koasati  in \sref{work} illustrates the
interplay of the various components in an implemented analysis, before
some conclusions are drawn in section \sref{conc}. 
\section{\label{red}Reduplication}
A well-known case from the context-sensitivity debate of the eighties is the
N-o-N reduplicative construction from Bambara (Northwestern Mande,
\cite{culy:85}):
\begin{examples}
\item \label{bambara}
\begin{tabular}[t]{@{}ll@{}}
a. & wulu-o-{\bf wulu} `whichever dog' \\
b. & wulunyinina-o-{\bf wulunyinina} \\ & `whichever dog searcher'\\
c. & wulunyininafil\`{e}la-o-{\bf wulunyininafil\`{e}la} \\ & `whoever watches dog
searchers'
\end{tabular}
\end{examples}
Beyond  total copying, \ref{bambara} also illustrates the
possibility of so-called fixed-melody parts in reduplication: a constant
/o/ intervenes between base (i.e.\/ original) and reduplicant
(i.e.\/ copied part, in bold print).%
\footnote{\newcite{culy:85}, who presents a superset of the data under 
  \ref{bambara} in the context of a formal proof of context-sensitivity, shows that the
  reduplicative construction in fact can copy the outcome of a {\em recursive}
  agentive construction, thereby becoming truly unbounded. He
   emphasizes the fact that it is ``very productive, with few, if any
   restrictions on the choice of the noun'' (
p.346).}

The next case from Semai expressive minor reduplication (Mon-Khmer, Hendricks
\shortcite{hendricks:98}) highlights the possibility of an interaction 
between reduplication and internal truncation:
\begin{examples}
\item \label{semai}
\begin{tabular}[t]{@{}llll@{}}
a. & c\textipa{PE:}t & {\bf ct}-c\textipa{PE:}t & `sweet' \\
b. & d\textipa{NO}h & {\bf dh}-d\textipa{NO}h & `appearance of nod-\\ & & & ding constantly' \\
c. & cfa\textlengthmark l & {\bf cl}-cfa\textlengthmark l & `appearance of flick-\\ & & & ering
red object'
\end{tabular}
\end{examples}
Reduplication copies the initial and final segment of the base,
skipping all of its interior segments, which may be of arbitrary length.

A final case comes from Koasati punctual-aspect reduplication (Muscogean, \cite{kimball:88}):
\begin{examples}
\item \label{koasati}
\begin{tabular}[t]{@{}lll@{}}
a. & ta.h\'as.pin & t$_1$ahas-{\bf t$_1$}\'o\textlengthmark-pin \\ & \multicolumn{2}{c}{`to be light in
weight'} \\
b. & la.p\'at.kin & l$_1$apat-{\bf  l$_1$}\'o\textlengthmark-kin \\ &  \multicolumn{2}{c}{`to be narrow} \\
c. & ak.l\'at.lin & a$_1$k-{\bf h$_1$}o-l\'atlin \\ &  \multicolumn{2}{c}{ `to be loose'} \\
d. & ok.c\'ak.kon & o$_1$k-{\bf h$_1$}o-c\'akkon \\ &  \multicolumn{2}{c}{`to be green or blue'} 
\end{tabular}
\end{examples}
Koasati is particularly interesting, because it shows that copy and
original need not always be adjacent  -- here the
reduplicant is infixed into its own base -- and also
because it illustrates that the copy may be phonologically modified:
the /h/ in the copied part of \ref{koasati}.c,d is best analysed as a voiceless vowel,
i.e.\/ the phonetically closest {\em consonantal} expression of its
source. Moreover, the locus of the infixed reduplicant is predictable
on prosodic grounds, as it is inserted after the first heavy syllable of
the base. Heavy syllables in Koasati are long %
(C)VV or closed (C)VC. Prosodic influence is also responsible for the length  
alternation of its fixed-melody part /o(o)/, since the heaviness
requirement for the penultimate, stressed, syllable of the word causes
long [o\textlengthmark] iff the reduplicant constitutes that syllable.

\section{\label{fsm}Finite-State Methods}
The present proposal differs from the state-labelled automata employed 
in One-Level Phonology by returning to conventional arc-labelled ones, but shares the idea that
labels denote {\em sets}, which is advantageous for compact automata.  
\subsection{Enriched Representations}
As motivated in \sref{red}, an appropriate automaton representation of
morphemes that may undergo reduplication should provide generic support for three key
operations: (i) copying or repetition of
symbols, (ii) truncation or skipping, and (iii) infixation.

For 
{\em copying}, the idea is to enrich the FSA representing a morpheme by
encoding stepwise repetition locally. For every content arc $ i
\stackrel{c}{\rightarrow} j$ we add a reverse {\bf repeat arc} $j
\stackrel{repeat}{\longrightarrow} i$. Following repeat arcs, we can
now move backwards within a string, as we shall see in more
detail below.

For 
{\em truncation}, a similar local encoding is available: For every content arc $ i
\stackrel{c}{\rightarrow} j$, add another  {\bf skip arc} $i
\stackrel{skip}{\longrightarrow} j$. This allows us to move forward
while suppressing the spellout of $c$.

A generic recipe for  {\em infixation} 
ensures that segmental material can be 
inserted anywhere within an existing morpheme FSA. A possible
representational enrichment therefore adds a
{\bf self loop} $i \stackrel{\Sigma}{\rightarrow} i$ labelled with the symbol
alphabet $\Sigma$ to every state $i$ of the FSA.%
\footnote{This can be seen as an application of the $ignore$ operator of
  \newcite{kaplan.kay:94}, where $\Sigma^{*}$ is being ignored.}

Each of the three enrichments presupposes an epsilon-free automaton in 
order to be wellbehaved. This requirement in particular ensures that
technical arcs ($skip$, $repeat$) are in 1:1 correspondence  with
content arcs, which is essential for unambiguous positional movement:
e.g.\/ $add\_skips(a\,\epsilon\,b)$ would ambiguously require 1 {\em or} 2
skips to supress the spellout of $b$, because it creates a disjunction of the empty string
$\epsilon$ with $skip$. It is perhaps worth
emphasizing that there is no special interpretation whatsoever for
these technical arcs: the standard automaton semantics is %
unaffected. As a consequence, $skip$ and $repeat$ will be a
visible part of the output in word form generation and must be allowed in the input for
parsing as well. 

\begin{figure}[htb]
\begin{tabular}{ll@{}}
\raisebox{2.5em}{a.} & \epsfig{file=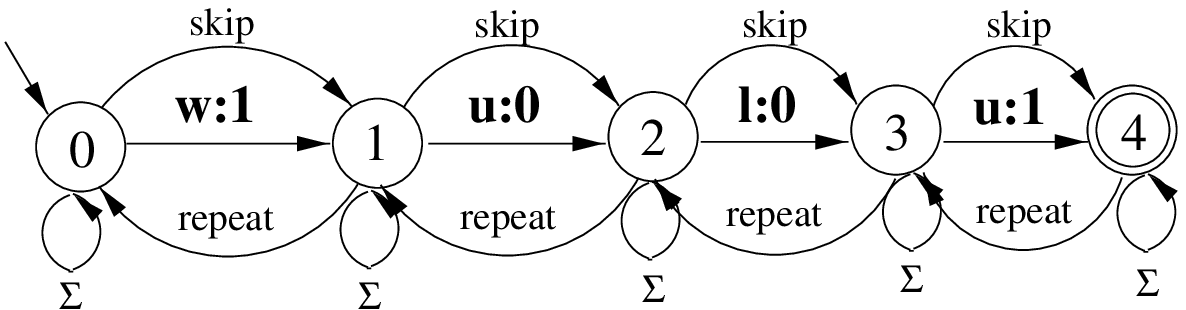,width=.85\linewidth} \\
\raisebox{2.0em}{b.} & \epsfig{file=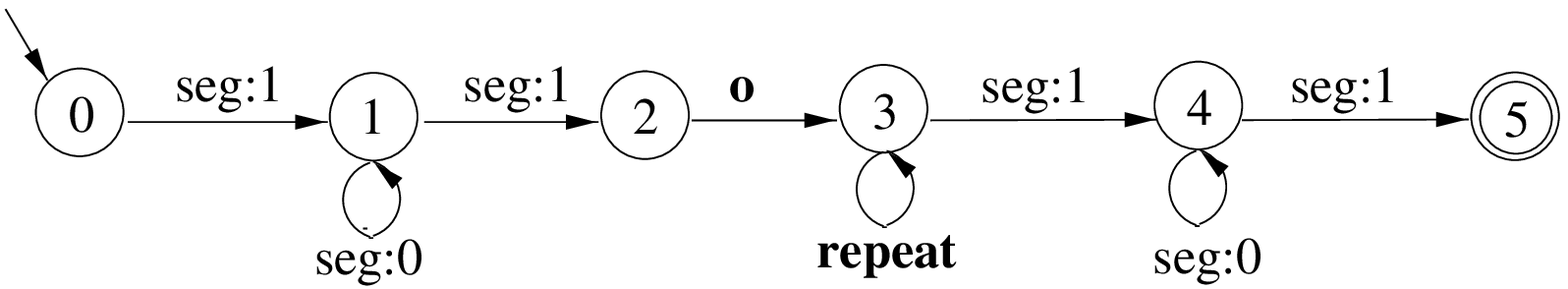,width=.85\linewidth} 
\end{tabular}
\caption{Enriched automata for \ling{wulu} (a.), Bambara N-o-N
  reduplication (b.)}\label{wulu} 
\end{figure}

Taken together, the three enrichments yield an automaton for Bambara \ling{wulu}, 
shown in figure \ref{wulu}.a. While skipping is not necessary for this
 example, $4 \stackrel{\Sigma}{\rightarrow} 4$ is: it will host the
fixed-melody /o/. The repeat arcs will of course facilitate copying, as we
shall see in a moment.
\subsection{Copying as Intersection}
Bird \& Ellison \shortcite{bird.ellison:92}  came close to discovering a useful device for
reduplication when they noted that automaton intersection has at least indexed-grammar
power (ibid., p.48). They demonstrated their claim by showing that
odd-length strings of indefinite length like the one described by the
regular expression $(a\, b\, c\, d\, e\, f\, g)^+$ can be repeated by intersecting them with an
automaton accepting only strings of even length: the result is  $(a\, b\, c\,
d\, e\, f\, g\, a\, b\, c\, d\, e\, f\, g)^+$.

Generalizing from their artifical example, let us first make one
additional minor enrichment by tagging the edges of the reduplicative
portion of a base with {\bf synchronization bits :1}, while using the
opposite value {\bf :0} for the interior part (see figure \ref{wulu}.a). This gives us a
segment-independent handle on those edges and a regular expression
$seg_{:1}{seg_{:0}}^{*}seg_{:1}$ for the whole synchronized portion ($seg$ abbreviates the
set of phonological segments).

Assuming repeat-enriched bases, a total reduplication
morpheme can now be seen as a partial word specification which mentions two
synchronized portions separated by an arbitrary-length move backwards: 
\begin{examples}
\item\label{total}
$seg_{:1}{seg_{:0}}^{*}seg_{:1}\;\boldsymbol{repeat}^{*}\;seg_{:1}{seg_{:0}}^{*}seg_{:1}$
\end{examples}
Moreover, total {\bf reduplicative copying} now simply {\bf  is  intersection}
of the base and \ref{total}, or -- in the Bambara case -- a simple
variant that adds the /o/ (figure \ref{wulu}.b). Disregarding self
loops for the moment, the reader may verify 
that no expansion of the kleene-starred $repeat$ that traverses less
than $|base|$ segments will satisfy the demand for {\em two}
synchronized portions. Semai requires another slight variant of \ref{total}
which {\em skips} the interior of the base in the reduplicant:
\begin{examples}
\item\label{semai_red}
$seg_{:1}\;\boldsymbol{skip}^{*}seg_{:1}\;repeat^{*}\;seg_{:1}{seg_{:0}}^{*}seg_{:1}$
\end{examples}
The identification of copying with intersection not only allows for
great flexibility in describing the full range of actual reduplicative
constructions with regular expressions, it also {\em reuses} the central
operation for constraint combination that is independently required
for one-level morphology and phonology. Any improvement in efficient
implementation of intersection therefore has immediate benefits for
grammar computation as a whole. In contrast, a hypothetical setup
where a dedicated total copy device is sandwiched between
finite-state transducers seems much less elegant and may require
additional machinery to detect copies during parsing. 

Note that it is in fact possible to compute reduplication-as-intersection
over an entire lexicon of bases (see figure
\ref{koasatiforms} for an example), provided that repeat arcs are
added individually to each base. Enriched base FSAs can then be
unioned together and undergo further automaton transformations such as 
determinization or minimization.
This restriction is necessary because our finite-state method cannot express token identity as
normally required in string repetition. Rather than identifying the
same token, it addresses the same string position, using the weaker
notion of type identity. Therefore, application of the method is only safe if strings
are effectively isolated from one another, which is exactly what
per-base enrichment achieves. See \sref{lazy} for a suggestion on how
to lift the restriction in practice.
\subsection{Resource Consciousness}
One pays a certain price for allowing general repetition and infixation: because of
its self loops and technical arcs, the automaton of figure
\ref{wulu}.a overgenerates wildly. Also, during intersection, self
loops can absorb other morphemes in unexpected ways. A possible
diagnosis of the underlying defect is that  we need to distinguish between {\bf producers and consumers of
  information}. In analogy to LFG's constraint vs constraining
equations, information may 
only be consumed if 
it has been produced at
least once. 

For automata, let us spend a P/C bit per arc, with P/C=1 for
producers and P/C=0 for consumer arcs. In {\em open interpretation} mode,
then, intersection combines the P/C bits of compatible arcs via
logical OR, making producers dominant. It follows that a resource may be
multiply consumed, which has obvious advantages for our application,
the multiple realization of string symbols. A final step of {\em closed
  interpretation} prunes all consumer-only arcs that survived
constraint interaction, in what may be seen as intersection with the universal
producer language under logical-AND combination of P/C bits. 

Using these resource-conscious notions, we can now model both the
default absence of material and purely contextual requirements as
consumer-type information: unless satisfied by lexical resources that
have been explicitly produced, the corresponding arcs will not be part
of the result. By convention, producers are displayed in bold. Thus, the exact
result of figure \ref{wulu}.a $\cap$ \ref{wulu}.b after closed interpretation is:
{\mathversion{bold}\[{w_{:1}\,u_{:0}\,l_{:0}\,u_{:0}\,o\,\,repeat^{4}\,repeat^{*}\,w_{:1}\,u_{:0}\,l_{:0}\,u_{:1}}\]} %
\noindent This expression also illustrates that, for parsing, strings like
\ling{wuluowulu} need to be consumer-self-loop-enriched via a small
preprocessing step, because intersection with the grammar would
otherwise fail due to unmentioned technical arcs such as
$repeat$. Because our proposal is fully declarative,
parsing then reduces to intersecting the enriched parse string with
the grammar-and-lexicon automaton (whose construction will itself involve
intersection) in closed interpretation mode,
followed by a check for nonemptiness of the result. Whereas the original parse
string was underspecified for morphological categories, the parse
result for a realistic morphology system will, in addition to
technical arcs, contain fully specified category arcs in some predefined
linearization order, which can be efficiently retrieved if desired. 
\subsection{\label{lazy}On-demand Algorithms}
It is clear that the above method is particularly attractive if some
of its operations can be performed online, since a fullform lexicon of 
productive reduplications  is clearly undesirable e.g.\/ for Bambara.
I therefore consider briefly questions of efficient implementation of
these operations.

Mohri et al.\/ \shortcite{mohri.pereira.riley:98} identify the
existence of a {\em local}
computation rule as the main precondition%
\footnote{A second condition is that no state is visited that has not
  been discovered from the start state. It is easy to implement
  \ref{loccomp} so that this condition is fulfilled as well.}
 for a {\em lazy} implementation of 
automaton operations, i.e.\/ one where results are only computed when demanded by
subsequent operations. Such implementations are very advantageous when 
large intermediate automata may be constructed but only a small part
of them is visited for any particular input. They show that such a rule exists for
composition $\circ$, hence also for our operation of intersection
 ($A \cap B \equiv range(identity(A) \circ identity(B))$).

Fortunately, the three enrichment steps all have local computation
rules as well:
\begin{examples}
\item \label{loccomp}
\begin{examples}
\item $q_1 \stackrel{c}{\rightarrow} q_2\;\Rightarrow\; q_2
\stackrel{repeat}{\longrightarrow} q_1$
\item $q_1 \stackrel{c}{\rightarrow} q_2\;\Rightarrow\; q_1
\stackrel{skip}{\longrightarrow} q_2$
\item $q \;\Rightarrow\; q \stackrel{\Sigma}{\rightarrow} q$
\end{examples}
\end{examples}
The impact of the %
existence of lazy implementations for
enrichment operations is twofold: we can (a) now maintain minimized base
lexicons for storage efficiency and add enrichments lazily {\em to the currently pursued
  string hypothesis} only, possibly modulated by exception diacritics
that control when enrichment should or should not happen.%
\footnote{See \newcite{walther:2000} for further details. With deterministic
  automata, the question of how to recover from a wrong string hypothesis during parsing
  is not an issue.}
And (b), laziness suffices to make the proposed reduplication method
reasonably time-efficient, despite the larger number of online operations.
Actual benchmarks from a pilot implementation are reported elsewhere \cite{walther:2000}.

\section{\label{work}A Worked Example}
In this section I show how to implement the Koasati case from
\ref{koasati} using the  FSA Utilities toolbox
\cite{vannoord:97}. FSA Utilities is a Prolog-based finite-state toolkit and
extendible regular expression compiler. It is freely available and encourages rapid prototyping.

Figure \ref{fsaops} displays the regular expression operators that
will be used (italicized operators are modifications or extensions).
\begin{figure}[htb]
\begin{tabular}[t]{cl}
\tt [] & empty string \\
\tt [E1,E2,\dots,En]    & concatenation of \tt E$_i$ \\
\tt \verb+{+E1,E2,\dots,En\verb+}+    & union of \tt E$_i$ \\
\tt E*                  & Kleene closure\\
\tt E\verb+^+                  & optionality\\
\tt E1 \verb+&+ E2      & intersection\\
\it X {-}{-}$>$ ( Y / Z) & monotonic rule \\ & $X \rightarrow Y\subseteq X\; /\;
\underline{\phantom{X}}\;Z$ \\
$\sim\; S$ &  complement set of $S$ \\
\it Head(arg1, \dots, argN) & (parametrized) \\  
 \it \phantom{Head(}:= Body &  macro definition 
\end{tabular}
\caption{Regular expression operators}\label{fsaops}
\end{figure}
The grammar will be presented below in a piecewise fashion, with line
numbers added for easy reference. 

Starting with the definition of
stems (line \l{1}), we %
add the three enrichments to  
the bare phonological string (\l{2}). However, the innermost
producer-type string constructed by \code{stringToAutomaton} (\l{3}) is
intersected with phonological constraints (\l{5,6}) that need to see the
string only, minus its enrichments. This is akin to lexical rule
application.

\noindent\begin{minipage}{\linewidth}
\startpiece\begin{verbatim}
1 stem(FirstSeg, String) := 
2  add_repeats(add_skips(add_self_loops(
3  [FirstSeg, stringToAutomaton(String)]
4    & ignore_technical_symbols_in(
5 moraification&mark_first_heavy_syllable
6    & positional_classification)))).
7
8 underspecified_for_voicing(BaseSpec) :=
9        { producer(BaseSpec & vowel), 
10         [producer(h),consumer(skip)] }.
11
12 tahaspin   := stem([], "tahaspin").
13 aklatlin   := stem(underspecified_for_
14               voicing(low),"klatlin").
\end{verbatim}\stoppiece\end{minipage}
Lines \l{8-10} capture the V/h alternation that is characteristic for vowel-initial stems 
under reduplication, with the vocalic alternant constituting the default used 
in isolated pronunciation. In contrast, the /h/ alternant is
concatenated with a \code{consumer}-type \code{skip} that requires a producer from
elsewhere. Lines \l{12-14} define two example stems. 

The following constraint (\l{15-18}) enriches a prosodically
underspecified string with moras $\mu$ -- abstract units of syllable weight 
\cite{hayes:95} --, a prerequisite to locating %
 (\l{20-24}) and synchronization-marking
(\l{25-31}) the first heavy syllable after which the reduplicative
infix will be inserted.\finalforcedpage
\startpiece\begin{verbatim}
15 moraification := 
16 ( vowel     --> ( mora / sigma ) )&
17 ( consonant --> ( mora / consonant ) )&
18 ( consonant --> ( (~ mora) / vowel ) ).
19
20 first_(X) := [not_contains(X), X].
21 heavy_rime := [consumer(mora),
22                consumer(mora)].
23 heavy_syllable := [consumer(~ mora), 
24                    heavy_rime].
25 mark_first_heavy_syllable := 
26 [first_(heavy_rime)&synced_constituent, 
27         synced_constituent].
28 right_synced := [consumer(~':1'&seg) *,
29                  consumer(':1'&seg)].
30 synced_constituent := 
31   [consumer(':1'&seg), right_synced].
32 positional_classification := 
33  [consumer(initial),consumer(medial) *, 
34   consumer(final)].
\end{verbatim}\stoppiece\finalpagebreak

\noindent Note that both the constituent before 
\smash{(\raisebox{.7em}{$\begin{smallmatrix}
        & \mu &    & \mu & \mu \\
        & |       &    & |      &  |      \\
t_{:1} & a     & h & a      &  s_{:1}
\end{smallmatrix}$})}
 and
{\em after}
(\raisebox{.7em}{$\begin{smallmatrix}
        & \mu   &  \mu \\
        & |         &   |      \\
p_{:1} & i      &   n_{:1}
\end{smallmatrix}$})
 the infixation site need to be marked.
Also, it turns out to be useful to classify base string positions for easy
reference in the reduplicative morpheme, which motivates lines \l{32-34}. 

The main part now is the reduplicative morpheme itself (\l{35}), which looks
like a mixture of Bambara and Semai: the spellout of the base
is followed by iterated repeats (\l{36}) to move back to its synchronized
initial position (\l{37}), which -- recall /h/  -- is required to be
consonantal. The rest of the base is skipped before insertion of the fixed-melody
part /o(o)/ occurs (\l{38, 42-44}). Proceeding with the interrupted
realization of the base, we identify its beginning as a synchronized
syllable onset ($\sim$ \code{mora}), followed by a right-synchronized
string (\l{39-40}).
\startpiece\begin{verbatim}
35 punctual_aspect_reduplication :=
36 [synced_constituent, producer(repeat)*,
37  consumer(':1' & initial & consonant), 
38  producer(skip) *, fixed_melody,
39  consumer(':1' & seg & ~ mora), 
40  right_synced].
41
42 fixed_melody := 
43 [producer(o & ~ ':1' & medial & mora),
44  producer(o & ~ ':1' & medial & mora)^].
\end{verbatim}\stoppiece

\begin{figure*}
   \centering
   \fbox{\begin{minipage}{\linewidth}
\epsfig{file=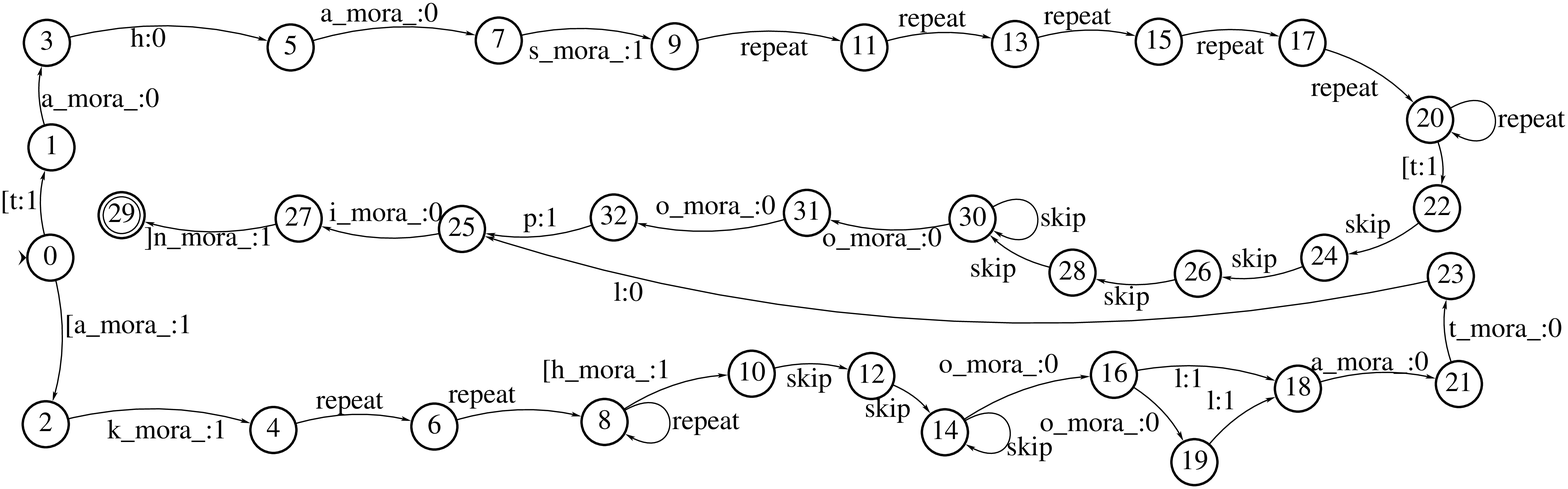,width=\linewidth}
\caption{Koasati reduplications \ling{tahas-too-pin}, \mbox{\ling{ak-ho(o)-latlin}}}\label{koasatiforms}
\end{minipage}}
\end{figure*}

Finally, some obvious \code{word\-\_level\-\_con\-straints} need to be defined
(\l{45-54}), before the central intersection of \code{Stem} and
punctual-aspect reduplication (\l{57}) completes our Koasati fragment: 
\startpiece\begin{verbatim}
45 word_level_constraints := 
46 last_segment_is_moraic & 
47 last_two_sylls_are_heavy.
48
49 last_segment_is_moraic := 
50   [consumer(sigma) *, consumer(mora)].
51
52 last_two_sylls_are_heavy := 
53   [consumer(sigma) *, 
54    heavy_syllable,heavy_syllable].
55
56 wordform(Stem):=closed_interpretation(
57   word_level_constraints & Stem & 
58   punctual_aspect_reduplication).
\end{verbatim}\stoppiece
The result of \code{wordform(\{tahaspin,aklatlin\})} is shown in figure
\ref{koasatiforms} (\code{[} and \code{]} are aliases for initial and final 
position).

Space precludes the description of a final automaton operation called Bounded
Local Optimization \cite{walther:99} that turns out to be
useful here to ban unattested free length variation, as found e.g.\/ in
\ling{ak-ho(o)-latlin} where the length of \ling{o} is yet to be determined.
Suffice to say that a parametrization of Bounded Local Optimization
would prune the moraic arc $16 \rightarrow 19$ in figure
\ref{koasatiforms} by considering it costlier than the non-moraic arc
$16 \rightarrow 18$, thereby eliminating the last  source of indeterminacy. 
\section{\label{conc}Conclusion}
This paper has presented a novel finite-state method for
reduplication that is applicable for both unbounded total cases, truncated or 
otherwise phonologically modified types and infixing instances. The
key ingredients of the proposal are suitably enriched automaton
representations, the identification of reduplicative copying with
automaton intersection and a resource-conscious interpretation that
differentiates between two types of arc symbols, namely producers and
consumers of information. After demonstrating the existence of
efficient on-demand algorithms to reduplication's central operations,
a case study from Koasati has shown that all of the above ingredients
may be necessary in the analysis of a single complex piece of prosodic
morphology. 
 
It is %
worth mentioning that our method can be transferred into a
two-level transducer setting without major difficulties \cite[appendix B]{walther:99}.

I conclude that the one-level approach  to reduplicative prosodic
morphology presents an attractive way of extending finite-state
techniques to difficult phenomena that hitherto resisted elegant computational analyses.
\section*{Acknowledgements}
The research in this paper has been funded by the German research
agency DFG under grant WI 853/4-1. Particular thanks go to the
anonymous reviewers for very useful comments.

\end{document}